\pdfoutput=1
\documentclass[11pt]{article}

\usepackage{naacl2021}

\usepackage{times}
\usepackage{latexsym}
\usepackage{hyperref}
\usepackage{url}
\usepackage{graphicx} % DO NOT CHANGE THIS
\usepackage{amsmath}
\usepackage{multirow}
\usepackage{multicol}
\usepackage{color}
\usepackage{booktabs}

\usepackage[T1]{fontenc}
\usepackage[utf8]{inputenc}

\usepackage{microtype}

\newenvironment{myitemize2}[1][]{%%%%%?????»???
\begin{list}{$\bullet$}
    {
     \setlength{\leftmargin}{5mm}     %?????
     \setlength{\parsep}{0.5mm}         %????????
     \setlength{\topsep}{0mm}         %??????????????????
     \setlength{\itemsep}{0mm}        %???????
     \setlength{\labelsep}{0.5em}     %?????????????????,???0.5em
     \setlength{\itemindent}{0mm}    %?????????
     \setlength{\listparindent}{6mm} %??????????
    }}
{\end{list}}%%%%%

\title{SemVLP: Vision-Language Pre-training by Aligning Semantics at Multiple Levels}

\author{Chenliang Li, Ming Yan, Haiyang Xu \\
\textbf{Fuli Luo, Wei Wang, Bin Bi, Songfang Huang} \\
  Alibaba Group \\
  {\tt \{lcl193798, ym119608, shuofeng.xhy\}@alibaba-inc.com} \\
  {\tt \{lfl259702, hebian.ww, b.bi, songfang.hsf\}@alibaba-inc.com}}

\begin{document}
\maketitle
\begin{abstract}
Vision-language pre-training (VLP) on large-scale image-text pairs has recently witnessed rapid progress for learning cross-modal representations. Existing pre-training methods either directly concatenate image representation and text representation at a feature level as input to a single-stream Transformer, or use a two-stream cross-modal Transformer to align the image-text representation at a high-level semantic space. In real-world image-text data, we observe that it is easy for some of the image-text pairs to align simple semantics on both modalities, while others may be related after higher-level abstraction. Therefore, in this paper, we propose a new pre-training method SemVLP, which jointly aligns both the low-level and high-level semantics between image and text representations. The model is pre-trained iteratively with two prevalent fashions: single-stream pre-training to align at a  fine-grained feature level and two-stream pre-training to align high-level semantics, by employing a shared Transformer network with a pluggable cross-modal attention module. An extensive set of experiments have been conducted on four well-established vision-language understanding tasks to demonstrate the effectiveness of the proposed SemVLP in aligning cross-modal representations towards different semantic granularities.
\end{abstract}

\section{Introduction}
\begin{figure*}
\centering
\includegraphics[width=0.97\textwidth]{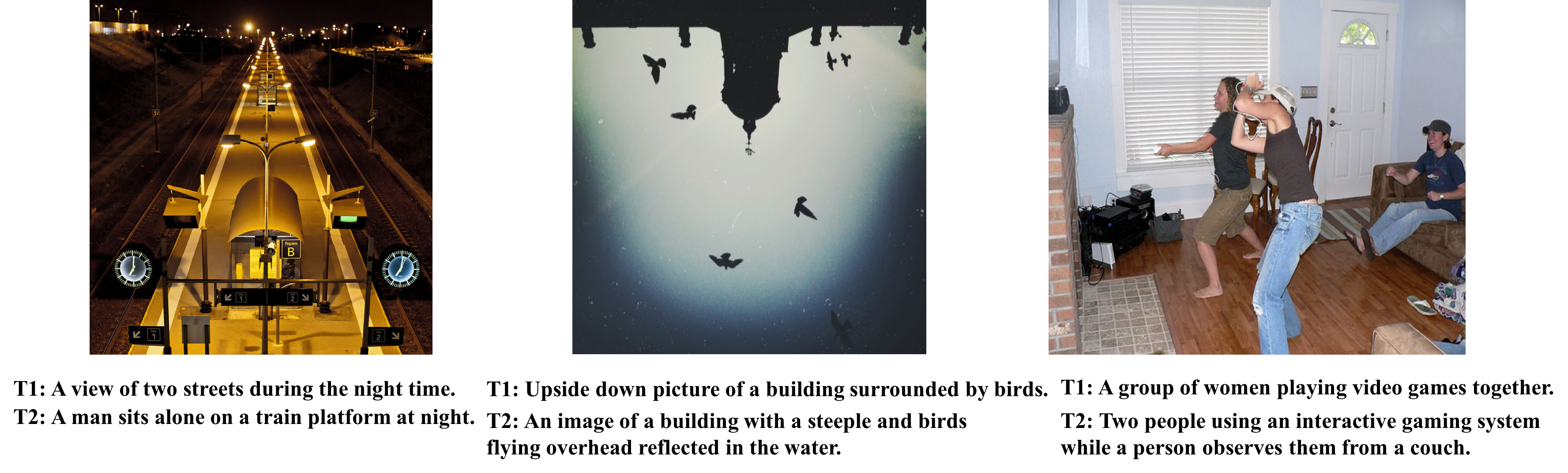}
\caption{Examples of images with two different caption text pieces from the MS COCO caption dataset, where some captions are more fine-grained than the others that are more abstract.} 
\label{fig:example} \vspace{-3mm}
\end{figure*}

Inspired by recent development of pre-trained language models in various NLP tasks, recent studies~\citep{lu2019vilbert,su2019vl,tan2019lxmert,chenuniter,li2020oscar,yu2020ernie} on vision-language pre-training (VLP) have pushed the limits of a variety of Vision-and-Language (V+L) tasks, which learn the semantic alignment between the different modalities by harnessing from large-scale image-text pairs. 
%\citep{lu2019vilbert,li2019visualbert,su2019vl,tan2019lxmert,chenuniter,li2020oscar,yu2020ernie}

The semantic gap between different modalities has always been treated as one of the most significant problems in cross-modality research. In current VLP literature, there are two mainstream architectures for bridging the cross-modal semantic gap: \textit{single-stream architecture} and \textit{two-stream architecture}. The former such as VL-BERT~\citep{su2019vl} and UNITER~\citep{chenuniter} assumes that the underlying semantics behind the two modalities is simple and clear, and thus simply concatenates image-region features and text features as input to a single Transformer~\citep{vaswani2017attention} network for early fusion in a straightforward manner. This paradigm learns the cross-modal semantic alignment from a bottom feature level by using the self-attention mechanism. Nevertheless, the design of single-stream structure treats both modality inputs equally, leaving the inherent different peculiarity of each modality not fully exploited. In contrast, the latter like LXMERT~\citep{tan2019lxmert} and ERNIE-ViL~\citep{yu2020ernie} first uses separate Transformer encoders to learn high-level abstraction of image and sentence representation respectively, and then combines the two modalities together with a cross-modal Transformer. This kind of design explicitly distinguishes between different modality inputs and aligns the cross-modal representations at a higher semantic level, but is usually parameter inefficient and may ignore certain more fundamental feature-level association. 

In real-world image-text data, we observe that it is easy for some of the image-text pairs to align simple semantics on both modalities, while others may be related after higher-level abstraction. As shown in Figure~\ref{fig:example}, the captions of T1 are more focused on the overview of the image with coarse-level semantics, while T2 are more detailed descriptions that emphasize on the specific parts of the images. The semantic granularity spans different levels for different captions of the same images. It is essential to explicitly consider aligning semantics at multiple levels for deeply understanding the real-world image-text data.

In light of this observation, we propose a new VLP pre-training architecture SemVLP, as a fusion of single-stream and two-stream architectures, which jointly aligns the image and text representation at multiple semantic levels. We observe that both the single-stream and two-stream architectures use common Transformer module, with the main difference that the latter introduces an extra CrossAttention module to allow cross-modal alignment at higher level and the different modalities are separately encoded. To complement the advantages of different architectures, we unify the two mainstream architectures by using a shared Transformer network and a pluggable cross-modal attention module, as shown in Figure~\ref{fig:framework}. To conduct more fine-grained feature-level alignment, we choose a single-stream mode and directly concatenate image-region features and text features as input to the shared Transformer network for pre-training. To enhance high-level semantic alignment, we switch to a two-stream mode by separately encoding both the image and text modalities with the same shared Transformer, where a cross-modal attention module is further added to allow cross-modal fusion at a higher semantic level. The pre-training procedure is conducted iteratively so as to align the real-world image-text data at multiple semantic levels. During the iterative pre-training phase, the shared Transformer network is forced to align the semantics at multiple levels, which enables the trained model to adapt to diverse image-text pairs. In this way, we take advantages of both mainstream architectures for cross-modal fusion, where the parameters are shared to allow for different pre-training styles that regularize with each other. 

%Specifically, different from the prevalent single-stream and two-stream Transformer architectures, we use a shared Transformer network with a pluggable cross-modal attention module for both the low-level and high-level semantic alignments, as shown in Figure~\ref{fig:framework}. For low-level semantic alignment, we directly concatenate image-region features and text features as input to the shared Transformer network for single-stream pre-training. For high-level semantic alignment, we introduce a novel two-stream Transformer network to align more abstract semantics by separately encoding the image and text parts with the shared Transformer, where a cross-modal attention module is further added to allow cross-modal fusion. The pre-training procedure is conducted iteratively to align the real-world image-text data at both semantic levels. During the iterative pre-training phase, the shared Transformer network is forced to align the semantics at multiple levels, which enables the trained model to adapt to diverse image-text pairs. In this way, we take advantages of both single-stream architecture and two-stream architecture for cross-modal fusion, where the parameters are shared to allow for different pre-training styles that regularize with each other. 

%~\citep{antol2015vqa}  %~\citep{suhr2018corpus} %~\citep{hudson2019gqa} %
We evaluate SemVLP on a variety of representative vision-language understanding tasks, including visual question answering, natural language visual reasoning and image-text/text-image retrieval. On all these tasks, SemVLP obtains significant improvements compared to those methods that align semantics at a single fixed level, where the proposed 12-layer SemVLP model outperforms all the previous single-stream and two-stream architectures with the same model size. 

%To demonstrate the effectiveness of SemVLP, we evaluate it on various of vision-language tasks, (1) visual question answering (VQA 2.0~\citep{antol2015vqa}), (2) natural language visual reasoning (NLVR2~\citep{suhr2018corpus}), (3) visual reasoning in the real world (GQA~\citep{hudson2019gqa}), and (4) image-text/text-image retrieval (Flickr30K~\citep{young2014image}). On all these tasks, SemVLP obtains significant improvements compared to those methods that align semantics at a single fixed level, where our 12-layer SemVLP model outperforms all the previous single-stream and two-stream architectures with the same model size. 

The main contributions of this work can be summarized as follows: (i) We introduce SemVLP, a simple and effective VLP method to learn generic image-text representations for V+L understanding tasks. (ii) We propose a new pre-training framework that aligns cross-modal semantics at multiple levels, which can take advantages of both single-stream and two-stream architectures. To the best of our knowledge, we are among the first to think about unifying the two mainstream architectures for better aligning the cross-modal semantics. 
(iii) We present extensive experiments and analysis to validate the effectiveness of the proposed SemVLP model, which can obtain superior performance with a 12-layer Transformer backbone on four V+L understanding tasks.

\section{Related Work}
Pre-training methods have substantially advanced the NLP field in both text understanding and text generation, such as BERT~\citep{devlin2018bert}, ALBERT~\cite{lan2019albert}, GPT~\citep{radford2018improving} and T5~\citep{raffel2019exploring}. Inspired by language pre-training, the research community starts to pay more attention to vision-language pre-training in multi-modal scenario, and many pre-training methods have been successfully applied to V+L tasks such as visual question answering~\citep{antol2015vqa,hudson2019gqa} and cross-modality retrieval~\citep{young2014image,lin2014microsoft}. In terms of the model architecture, there are mainly two broad directions to conduct vision-language pre-training. The first line uses a single-stream transformer architecture to model both image and text representations in a unified semantic space such as VLBERT~\citep{su2019vl}, UNITER~\citep{chenuniter} and OSCAR~\citep{li2020oscar}. In contrast, the other line adopts a two-stream Transformer architecture that first encodes the image and text modalities separately, and then fuses the cross-modal representations with another Transformer network. Furthermore, some other works focus on designing different pre-training tasks to learn better cross-modal representations such as ERNIE-ViL~\citep{yu2020ernie} and PixelBERT~\citep{huang2020pixel}.

In this paper, we focus on the V+L understanding tasks with VLP method. Instead of choosing only one model architecture for VL pre-training, we introduce a pioneer work of fusing both the single-stream and two-stream architectures to better align the cross-modal semantics at multiple levels. 

\section{SemVLP Pre-training}
%In this section, we will first introduce the model architecture of SemVLP. Then we will illustrate how we align cross-modal semantics at multiple levels with SemVLP model. Finally, the pre-training tasks and strategy will be introduced.

\subsection{Model Architecture}
\begin{figure*}
\centering
\includegraphics[width=0.92\textwidth]{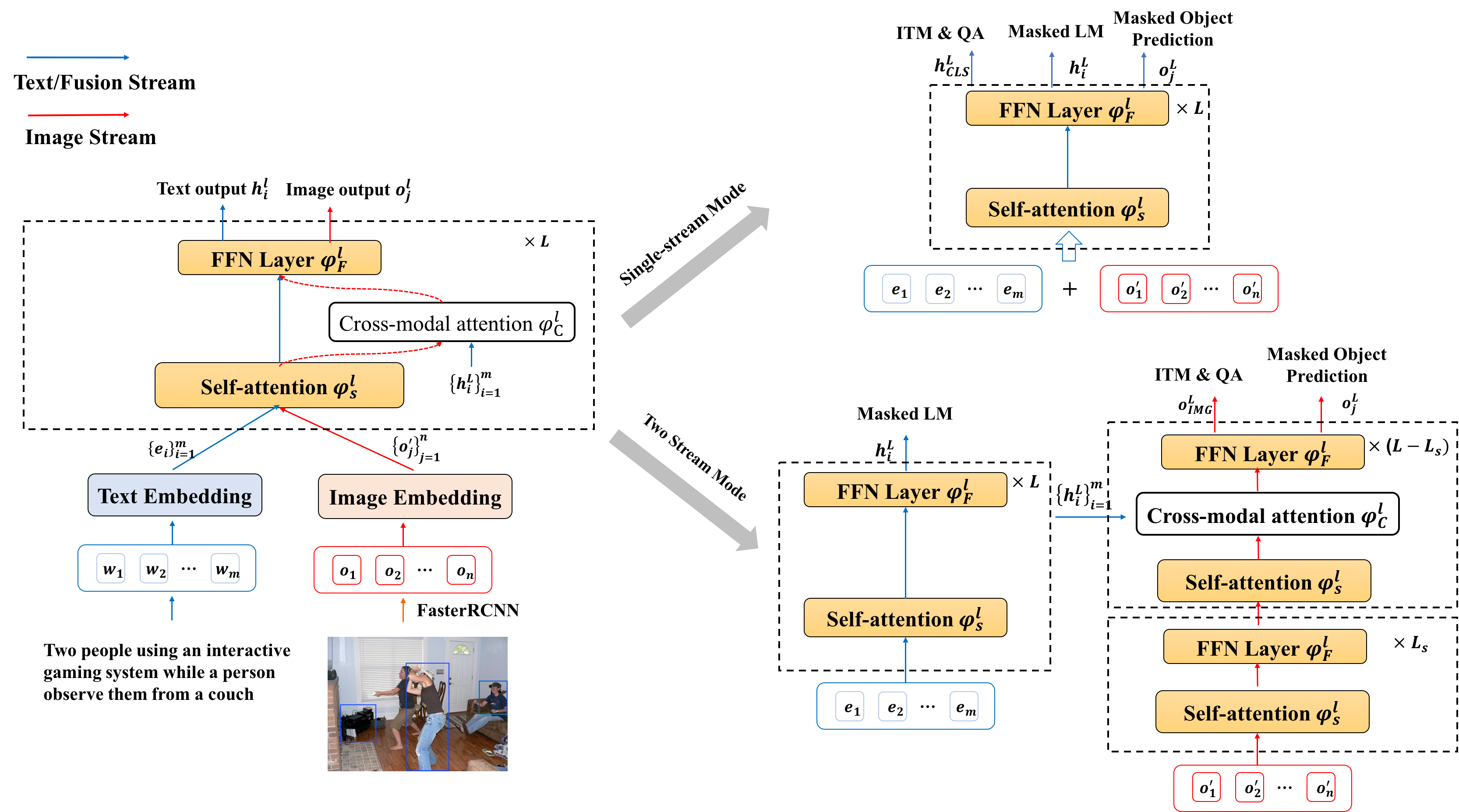}
\caption{The overall framework of SemVLP. The whole model consists of a shared multi-layer Transformer encoder and an extra cross-modal attention module $\varphi_C^l$, where the Transformer encoder includes a self-attention module $\varphi_S^l$ and a FFN layer $\varphi_F^l$ (feed-forward network). In single-stream mode, we directly concatenate the image and text representation as input to the shared Transformer encoder for V+L pre-training. In two-stream mode, we reuse the same shared Transformer encoder to separately encode both the image and text representations, and add an extra cross-modal attention to align the semantics at a higher level. 
%The pre-training is conducted with an iterative training schedule. 
} 
%\caption{The overall framework of SemVLP. We use a shared Transformer network and a pluggable cross-modal attention module to support both the low-level and high-level semantic alignments in an iterative learning framework.} 
\label{fig:framework} \vspace{-2mm}
\end{figure*}

The architecture overview of SemVLP is shown in Figure~\ref{fig:framework}. Inspired by the idea of sharing the encoder and decoder in Transformers for neural machine translation~\citep{xia2019tied}, we base the architecture of SemVLP on a shared bidirectional Transformer encoder, where a pluggable cross-modal attention module is further used to enhance high-level semantic alignment. By sharing the model parameters and adjusting the input format, SemVLP can be flexible to switch between single-stream and two-stream pre-training architectures, with the input text and image encoded in different semantic levels. In this way, we cast both the mainstream pre-training architectures into a more compact one in that there is only one copy of parameter set, which is applicable to both the low-level and high-level semantic alignment with much less parameter cost. We iteratively pre-train on the two settings towards better understanding of the real-world image-text pairs.

\subsubsection{Input Embeddings}
The input to SemVLP is an image and its related sentence (e.g. caption text). Each image is represented as a sequence of objects $\{o_{1},...,o_{n}\}$, and each sentence is represented as a sequence of words $\{w_{1},...,w_{m}\}$. After cross-modal fusion and alignment at multiple semantic levels, SemVLP is able to generate language representations, image representations and cross-modal representations from the image-text inputs. Given the sequence of words and objects, we first introduce the methods to embed the inputs to the feature space. 

\textbf{Sentence Embeddings} \quad We adopt the same method as BERT~\citep{devlin2018bert}, which uses WordPiece tokenizer to tokenize the input sentence into sub-word tokens. The sequence of input tokens is as $\{[CLS], w_{1},...,w_{m}, [SEP]\}$, where $[CLS]$ and $[SEP]$ are special tokens in BERT. The final embedding $e_i$ for each token is generated by combining the original word embedding, segment embedding and position embedding. 

\textbf{Image Embeddings} \quad We use a pre-trained object detector Faster R-CNN~\citep{ren2015faster} to extract the object-level image features from the image, where each object $o_j$ is represented as a 2048-dimensional feature vector $f_j$. To capture the spatial information of the object, we also encode the box-level location features for each object via a 4-dimensional vector $l_j = (\frac{x_1}{W},\frac{y_1}{H},\frac{x_2}{W},\frac{y_2}{H})$, where $(x_1, y_1)$ and $(x_2, y_2)$ denote the coordinate of the bottom-left and top-right corner while $W$ and $H$ are the width and height of the input image. We concatenate $f_j$ and $l_j$ to form a position-sensitive object feature vector, which is further transformed into $o'_j$ using a linear projection to ensure that it has the same vector dimension as that of word embeddings. Similar to special token $[CLS]$ in sentence embeddings, we also add a special feature $[IMG]$ to denote the representation of the entire image and add it to the beginning of the input object sequence. 

\subsection{Shared Transformer Encoder}
% we use a shared bidirectional Transformer encoder with a pluggable cross-modal attention module to better learn the cross-modal representations at multiple semantic levels, as shown in Figure~\ref{fig:framework}. 
Given the embeddings of the words for the sentence $\{e_i\}_{i=1}^m$ and the image regions $\{o'_j\}_{j=1}^n$, the full encoder is a stacked model with $L$ blocks, where the $l'$th block consists of a self-attention module $\varphi_S^l$, a nonlinear feed forward network $\varphi_F^l$ and a pluggable cross-modal attention module $\varphi_C^l$, where superscript $l$ represents the layer index. Both the self-attention and cross-modal attention modules are based on the multi-head attention~\citep{vaswani2017attention}, where the feed forward network (FFN) consists of an intermediate layer and an output layer as in BERT~\citep{devlin2018bert}. In both the single-stream and two-stream modes, the self-attention module $\varphi_S^l$ and feed forward network $\varphi_F^l$ are shared and tied to a single one copy of parameter set. The cross-modal attention module $\varphi_C^l$ is additionally used in two-stream mode to enhance the high-level semantic alignment. 
%Next we will introduce the details of our method by combining different encoder modules to align the cross-modal representations at multiple semantic levels.

\subsubsection{Feature-level Semantic Alignment}
%$S^0 = \{o'_1,o'_2,...,o'_n,e_1,e_2,...,e_m\}$
To allow a fine-grained feature-level semantic alignment, we directly concatenate the image and text embedding features as input to the single-stream mode of SemVLP, which consists of the shared self-attention module and nonlinear FFN layer. Specifically, we initialize $S^0 = \{o'_1,...,o'_n,e_1,...,e_m\}$. The encoding process can be formulated as: 
\begin{equation*}
\begin{aligned}
s_i^l &= \varphi_F^l(\varphi_S^l(s_i^{l-1}, S^{l-1})) \\
S^l&=\{s_1^l, s_2^l, ..., s_{n+m}^l\}=\{o^{l}_1,...,o^{l}_n,h^{l}_1,...,h^{l}_m\} 
%S^l&=\{s_1^l, s_2^l, ..., s_{n+m}^l\}=\{o^{l}_1,o^{l}_2,...,o^{l}_n,h^{l}_1,h^{l}_2,...,h^{l}_m\} 
\end{aligned}
\end{equation*}
where $\{h^l_i\}$ and $\{o^l_j\}$ are the text and object representation of layer $l$, respectively. In this way, we can get full interaction between the image and text representations from a bottom feature-level embedding space. Eventually, we obtain $O^L=\{o^{L}_1,o^{L}_2,...,o^{L}_n\}$ and $H^L=\{h^{L}_1,h^{L}_2,...,h^{L}_m\}$, the representations of all the object outputs and text outputs of the last layer in the SemVLP encoder. The hidden representations $O^L$ and $H^L$ are then used to conduct the subsequent pre-training tasks.

\subsubsection{High-level Semantic Alignment}
For enhancing high-level semantic alignment, we adopt the two-stream mode of SemVLP, where text and image objects are separately encoded first and then fuse at a high-level semantic space. Therefore, we adopt a two-encoder architecture shown on bottom-right of Figure~\ref{fig:framework}. The two-stream design of SemVLP mainly derives from the Transformer encoder-decoder network, where the main difference lies in: (1) we use both the bidirectional Transformer encoders for encoding the image and text inputs, which focuses on the V+L understanding tasks, (2) except for the cross-modal attention module, the image and text encoders share the same model parameters. We find that such parameter sharing can enhance the semantic alignment at a module level, and act as a form of regularization that stabilizes the training and saves memory consumption~\citep{xia2019tied}, (3) different from previous Transformer encoder-decoder architecture which introduces the cross-attention module to all blocks of the decoder, we only introduce the cross-modal attention module at the upper parts of the blocks, so as to better fuse the cross-modal representations at high-level semantic space. The encoding process of two-stream mode can be formulated as follows: 
%It consists of the shared self-attention module, cross-modal attention module and nonlinear FFN layer. To make it possible to separately encode the text and image representations with the SemVLP model, we adopt a two-encoder architecture shown in the bottom-right part of Figure~\ref{fig:framework} by tying all the parameters of self-attention module and FFN layer of the text encoder and image encoder, where a cross-modal attention module is further used to fuse the cross-modal representations. Different from the previous Transformer encoder-decoder architecture which introduces the cross-attention module to all blocks of the decoder, we only introduce the cross-modal attention module at the upper parts of the blocks, so as to better fuse the cross-modal representations at high-level semantic space. Specifically, we initialize $H^0 = \{e_1,e_2,...,e_m\}$ and $O^0 = \{o'_1,o'_2,...,o'_n\}$. The encoding process of two-stream mode can be formulated as follows: 
\begin{equation*}
\begin{aligned}
h_j^l &= \varphi_F^l(\varphi_S^l(h_j^{l-1}, H^{l-1})) \\
o_j^l &= \varphi_F^l(\varphi_S^l(o_j^{l-1}, O^{l-1})),  s.t. \quad l <= L_s \\
o_{j+1}^l &= \varphi_F^l(\varphi_C^l(\varphi_S^l(o_{j+1}^{l-1}, O^{l-1}), H^L)),  s.t. \quad l > L_s 
\end{aligned}
\end{equation*}
where $L_s$ indicates the layer index that cross-modal attention is introduced. The image and text feature embeddings are first separately encoded, and then the hidden states $H^L$ of text output in the last layer are used as input to the cross-modal attention for better understanding the image representations. Eventually, we can obtain the output representations of image objects and text, $O^L=\{o^{L}_1,o^{L}_2,...,o^{L}_n\}$ and $H^L=\{h^{L}_1,h^{L}_2,...,h^{L}_m\}$. With $O^L$ and $H^L$, we could use a simple network with a softmax layer to conduct the subsequent pre-training tasks. 

\subsection{Joint Training}
%In this section, we first introduce the pre-training tasks used in our method, then the training strategy in terms of different semantic alignments will be introduced. 

\subsubsection{Pre-training Tasks}
We follow LXMERT~\citep{tan2019lxmert} and use three-types of pre-training tasks: i.e., language task, vision task and cross-modality task. 

\textbf{Masked LM Prediction} \quad The task setup is basically the same as in BERT~\citep{devlin2018bert}, we randomly mask 15\% tokens in the text and the model is asked to predict these masked words with the output text representations $H^L$. For different pre-training modes, the masked words will be predicted either with the help of visual modality so as to resolve ambiguity (single-stream mode), or from text modality alone so as to increase task difficulty (two-stream mode). 

\textbf{Masked Object Prediction} \quad Similarly, we pretrain the vision side by randomly masking objects, i.e., the object features are masked with zeros. We randomly mask 15\% image objects and ask the model to predict properties of these masked objects with the output object representations $O^L$. To capture more object-level semantics, we follow the object prediction task in LXMERT~\citep{tan2019lxmert} and perform two sub-tasks: ROI-Feature Regression and Detected Label Classification. We take the detected labels output by Faster R-CNN~\citep{ren2015faster} as the ground-truth labels for prediction. 

\textbf{Image-Text Matching (ITM)} \quad The task setup is almost the same as in LXMERT~\citep{tan2019lxmert}, that we randomly sample 50\% mismatched image-text pairs and 50\% matched pairs, and train an classifier to predict whether an image and a sentence match each other on the representation $\textbf{h}^L_{CLS}$ (single-stream mode) and $\textbf{o}^L_{IMG}$ (two-stream mode). One difference is that we do not enforce the masked LM prediction and Object Prediction loss when sampling a mismatched image-text pair. 

\textbf{Image Question Answering (QA) } \quad We also cast the image question answering task as a classification problem and pre-train the model with image QA data as in LXMERT~\citep{tan2019lxmert}, which leads to a better cross-modality representation. We build the classifier on top of the representation $\textbf{h}^L_{CLS}$ for single-stream mode and on that of $\textbf{o}^L_{IMG}$ for two-stream mode.

\subsubsection{Pre-training Strategy}

SemVLP is pre-trained with multiple pre-training tasks and we add all these task losses with equal weights. To jointly align semantics at multiple levels, given a mini-batch of image-text pairs, 50\% of the time we update the model with single-stream mode, while 50\% of the time we update it with two-stream mode. In this way, for every update of SemVLP, the model is pre-trained at multiple semantic levels, so as to better model the diverse image-text data. 

\subsubsection{Fine-tuning Ingredient}

After pre-training is completed, SemVLP can support fine-tuning with either a single-stream architecture or a two-stream architecture. In single-stream mode, the hidden state $h^L_{CLS}$ of the last layer is used for cross-modality calculation, while the hidden state of $o^L_{IMG}$ is used in two-stream mode. For each downstream task, we examine the performances for both the single-stream and two-stream fine-tuning. To yield a single model result, we use only the architecture mode with the optimal performance on development set for final evaluation.

\section{Experiments}
\begin{table*}
\centering
\small
\begin{tabular}{cl|c|ll|lll|lll}
\toprule
\multicolumn{2}{c|}{\multirow{2}{*}{Models}}      &
\multirow{2}{*}{Params} &
\multicolumn{2}{c|}{VQA} & \multicolumn{3}{c|}{IR-Flickr30K} & \multicolumn{3}{c}{TR-Flickr30K}  \\
\multicolumn{2}{l|}{}                 &            & Test-dev & Test-std     & R@1   & R@5   & R@10             & R@1   & R@5   & R@10              \\
\midrule
\multirow{5}{*}{Single-stream} & VisualBERT &  110M    & 70.80    & 71.00        & -     & -     & -                & -     & -     & -                 \\
                  & VLBERT & 110M   & 71.16    & -        & - & - & -            & - & - & -             \\
			      & Unicoder-VL & 110M & - & - & 71.50 & 90.90 & 94.90 & 86.20 & 96.30 & 99.00 \\
                               & UNITER & 110M     & 72.70    & 72.91        & 72.52 & 92.36 & 96.08            & 85.90 & 97.10 & 98.80             \\
                               & OSCAR & 110M       & 73.16    & 73.61        & -     & -     & -                & -     & -     & -                 \\
%			      & PixelBERT-x152 & 227M  & 74.45    & 74.55        & 71.50 & 92.10 & 95.80            & 87.00 & \textbf{98.90} & \textbf{99.50}             \\
\midrule
\multirow{4}{*}{Two-stream}    & ViLBERT & 221M & 70.55    & 70.92        & 58.20 & 84.90 & 91.52            & -     & -     & -                 \\
                               & 12-in-1 & 221M & 73.15    & -        & 67.90 & - & -            & -     & -     & -                 \\
                               & LXMERT & 183M          & 72.42    & 72.54        & -     & -     & -                & -     & -     & -                 \\
                               & ERNIE-ViL & ~210M  & 72.62    & 72.85        & 74.44 & 92.72 & 95.94            & 86.70 & 97.80 & 99.00             \\
\midrule
Our  Model             & SemVLP & 110M/140M  & \textbf{74.52}    & \textbf{74.68}        &     \textbf{74.80}  & \textbf{93.43}      &   \textbf{96.12}               &    \textbf{87.70}   &   \textbf{98.20}    &      \textbf{99.30}            \\
\bottomrule
\end{tabular}
%& PixelBERT-r50   & 71.35    & 71.42        & 59.80 & 85.50 & 91.60            & 75.70 & 94.70 & 97.10             \\
%& PixelBERT-x152  & 74.45    & 74.55        & 71.50 & 92.10 & 95.80            & 87.00 & \textbf{98.90} & \textbf{99.50}             \\
%& VILLA-base   & 73.59    & 73.67        & 74.74 & 92.86 & 95.82            & 75.70 & 94.70 & 97.10             \\
%& VILLA-large  & 74.69    & 74.87        & 76.26 & 94.24 & 96.84            & 87.90 & 97.50 & 98.80             \\
\caption{Evaluation Results on VQA and Flickr30K.}
\label{table:overall1}
\end{table*}
\subsection{Pre-training Setup}
\textbf{Pre-training Data} \quad We use the same in-domain data as in LXMERT~\citep{tan2019lxmert} for pre-training. It consists of the image caption data from MS COCO~\citep{lin2014microsoft}, Visual Genome~\citep{krishna2017visual}, and image question answering data from VQA v2.0~\citep{antol2015vqa}, GQA balanced version~\citep{hudson2019gqa} and VG-QA~\citep{zhu2016visual7w}. The total amount of the dataset is 9.18M image-and-sentence pairs on 180K distinct images. Besides, we also use additional out-of-domain data from Conceptual Captions~\citep{sharma2018conceptual} and SBU Captions~\citep{ordonez2011im2text} for model pre-training, which consists of about 4M image-text pairs on 4M images. 
%and SBU Captions~\citep{ordonez2011im2text}

\textbf{Implementation Details} \quad The maximum sequence length for the sentence is set as 20. We use Faster R-CNN~\citep{ren2015faster} (with ResNet-101 backbone~\citep{he2016deep}) pre-trained on Visual Genome dataset~\citep{krishna2017visual} to detect the objects and extract the region features. We consistently keep 100 objects for each image to maximize the pre-training compute utilization by avoiding padding. For the model architecture, we pre-train a 12-layer SemVLP-base model with hidden size of 768, where we initialize it with the parameters from StructBERT base model~\citep{wang2019structbert}. We set $L_s=6$, which obtains the best performances on the development set of the downstream tasks, at a proper semantic level for cross-modal fusion~\footnote{\small{We only introduce the cross-modal attention from text space to image space due to the superior performance in our framework, where modeling of vision modality is emphasized.}}. We train SemVLP model with a total batch size of 256 for 40 epochs on 4 V100 GPUs. The Adam optimizer with initial learning rate of 1e-4 and a learning rate linear decay schedule is utilized.

\begin{table*}
\centering
\small
\begin{tabular}{l|ccccc|c} 
\toprule
\multirow{2}{*}{Models} & MMN & NSM   & \multirow{2}{*}{LXMERT} & \multirow{2}{*}{12-in-1}  & \multirow{2}{*}{OSCAR} & \multirow{2}{*}{SemVLP}  \\
                        &  \citep{chen2019meta}    &   \citep{hudson2019learning}    &        &       &       &              \\ 
\midrule
Test-dev                & -  & -     & 60.00       & -     & 61.58 & \textbf{62.87}        \\
Test-std                & 60.83  & 63.17 &  60.33   & 60.65 & 61.62 & \textbf{63.62}        \\
\bottomrule
\end{tabular}
\caption{Evaluation Results on GQA.}
\label{table:overall2}
\end{table*}

\begin{table*} [!htb]
\centering
\small
\begin{tabular}{c|cccc|c} 
\toprule
Models (params) & VisualBERT(110M) & LXMERT(183M) & UNITER(86M) & OSCAR(110M) & SemVLP(110M/140M)  \\
\midrule
Dev    & 67.40      & 74.90  & 77.14               & 78.07      & \textbf{79.00}        \\
Test-P  & 67.00      & 74.50  & 77.87               & 78.36      & \textbf{79.55}        \\
\bottomrule
\end{tabular}
\caption{Evaluation Results on NLVR2.}
\label{table:overall3}
\end{table*}

\subsection{Results on Downstream Tasks}
%our pre-training 
We compare SemVLP model against other state-of-the-art single-stream and two-stream cross-modal pre-training models of the comparable model size~\footnote{\small{Most of the compared models have similar model size as 12-layer BERT-base.}} on the following downstream tasks. The details on these tasks and fine-tuning configurations can be found in the supplementary material.

%\begin{enumerate}
%\itemsep0em
%\item \textbf{VQA v2.0}~\cite{antol2015vqa}: A visual question answering task/dataset that asks a model natural language questions on a given image.
%\item \textbf{Image-Text Retrieval}: We conduct experiments on the Flickr30K dataset~\citep{young2014image}.
%\item \textbf{NLVR2}~\cite{suhr2018corpus}: A visual reasoning task that aims to determine whether a natural language statement is true about a pair of images.
%\item \textbf{GQA 2019}~\cite{hudson2019gqa}: An image question answering task/dataset that emphasizes on the reasoning capability of a model to answer a question.
%\end{enumerate}

\begin{myitemize2}
\itemsep0em
\item \textbf{VQA v2.0}~\cite{antol2015vqa}: A visual question answering task/dataset that asks a model natural language questions on a given image.
\item \textbf{Image-Text Retrieval}: We test on the popular Flickr30K dataset~\citep{young2014image}.
\item \textbf{NLVR2}~\cite{suhr2018corpus}: A visual reasoning task that aims to determine whether a natural language statement is true about a pair of images.
\item \textbf{GQA 2019}~\cite{hudson2019gqa}: An image question answering task/dataset that emphasizes on the reasoning capability of a model to answer a question.
\end{myitemize2}

%To account for parameter efficiency, we also list  two types of models with different model sizes: (i) the VLP models of similar size to BERT base. (ii) the VLP models that have similar size to BERT large. 
The results on the four downstream V+L tasks are shown in Table~\ref{table:overall1},\ref{table:overall2},\ref{table:overall3} respectively. We can see that: (1) Among all the VLP models of similar size to BERT base, SemVLP consistently outperforms other strong single-stream and two-stream VLP methods (e.g., UNITER~\citep{chenuniter}, OSCAR~\citep{li2020oscar} and 12-in-1~\citep{lu202012}, ERNIE-ViL~\citep{yu2020ernie}) on all the examined tasks, which validates the effectiveness of SemVLP on combining single-stream and two-stream architectures to align semantics at multiple levels. (2) With much less parameters, the single-stream architecture can achieve comparable performance to the two-stream architecture, which is more parameter efficient. The proposed SemVLP model can be easily adapted to either architecture according to the typical scenario. By sharing parameters, SemVLP can also be parameter efficient while keeping superior performance. It is partially because SemVLP is pre-trained to align cross-modal semantics  at multiple semantic levels, which makes the learning of semantic alignments more robust toward the diverse image-text pairs. 

%the proposed SemVLP model  using a 12-layer base backbone can achieve performances comparable to those of the previous state-of-the-art methods on almost all the tasks, and even outperform many large VLP models (e.g.,  VLBERT-large~\citep{su2019vl} and UNITER-large~\citep{chenuniter}) on VQA and NLVR2 tasks. In all the VLP models of similar size to BERT base, our SemVLP model consistently outperforms other strong VLP base models (e.g., LXMERT~\citep{tan2019lxmert}, UNITER-base~\citep{chenuniter}) on most tasks, and often by a significantly large margin. It demonstrates that the proposed SemVLP model is highly parameter-efficient, partially because it is pre-trained to align cross-modal semantics  at multiple semantic levels, which makes the learning of semantic alignments more robust toward the diverse image-text pairs. 
\begin{figure}
\centering
\includegraphics[width=0.49\textwidth]{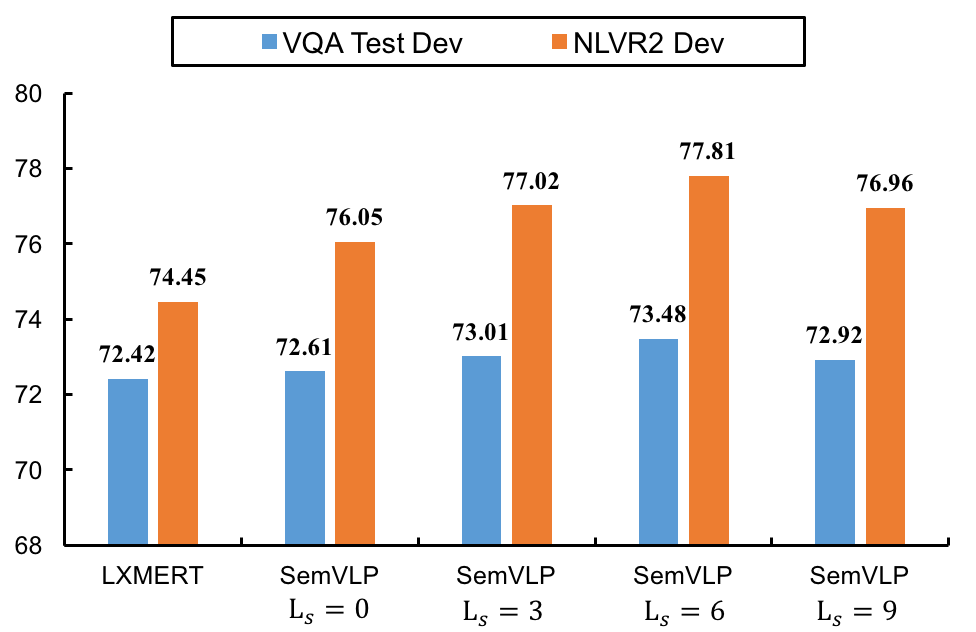}
\caption{Results w.r.t different two-stream architectures for aligning high-level semantics on VQA and NLVR2 development set.} 
%(best performance achieved at the semantic level of $L_s=6$)
\label{fig:analysis2} \vspace{-2mm}
\end{figure}
\subsection{Pre-training on Different Semantic Levels}
To validate the effectiveness of aligning cross-modal semantics at multiple levels, we conduct in-depth analysis on pre-training at different semantic levels with various architectures.

\textbf{Analysis on Various Pre-training Architectures} \quad We first examine the importance of pre-training at multiple semantic levels by conducting ablation study on the pre-training fashions. Specifically, we pre-train the SemVLP model with only one type of model architecture each time and test the performance on the downstream tasks. All the pre-training settings are kept the same as in the original SemVLP pre-training. As shown in Table~\ref{table:analysis1}, both the two pre-training fashions play important roles in obtaining the full SemVLP model, and removing each task will consistently decrease the final downstream task performance. The single-stream architecture is used to align fair-grained feature-level semantics, while the two-stream architecture helps align semantics at a higher-level semantic space. By iterative training with a shared set of Transformer parameters, the proposed SemVLP model can take the advantage of both the single-stream architecture and two-stream architecture towards more robust vision-language pre-training. 

\begin{table}
\centering
\small
\begin{tabular}{l|c|c|c} 
\toprule
%{\multirow{2}{*}{Models}}  & VQA & GQA & NLVR2 \\
%                            & Test-dev    & Test-dev & Dev             \\ 
Pre-training Mode  & VQA & GQA & NLVR2 \\
\midrule
Baseline 1 (single-stream) & 73.72 & 61.82    & 78.02 \\
Baseline 2 (two-stream) & 73.48 & 61.68    & 77.81 \\
SemVLP & \textbf{74.52} & \textbf{62.87}    & \textbf{79.00} \\
\bottomrule
\end{tabular}
\caption{Ablation study of different pre-training fashions on development set. Baseline 1 and Baseline 2 denote pre-training with only single-stream or only two-stream architecture, respectively.}
\label{table:analysis1}
\end{table}

\begin{table}
\centering
\small
\begin{tabular}{l|c|c|c} 
\toprule
Fine-tuning Mode  & VQA & GQA & NLVR2 \\
\midrule
Single-stream & \textbf{74.52} & \textbf{62.87}    & \textbf{79.00} \\
Two-stream & 73.92 & 62.18  & 78.48 \\
\bottomrule
\end{tabular}
\caption{Results w.r.t. different fine-tuning architectures after the SemVLP model is fully pre-trained.}
\label{table:finetune} \vspace{-2mm}
\end{table}

\begin{figure*}
\centering
\includegraphics[width=0.85\textwidth]{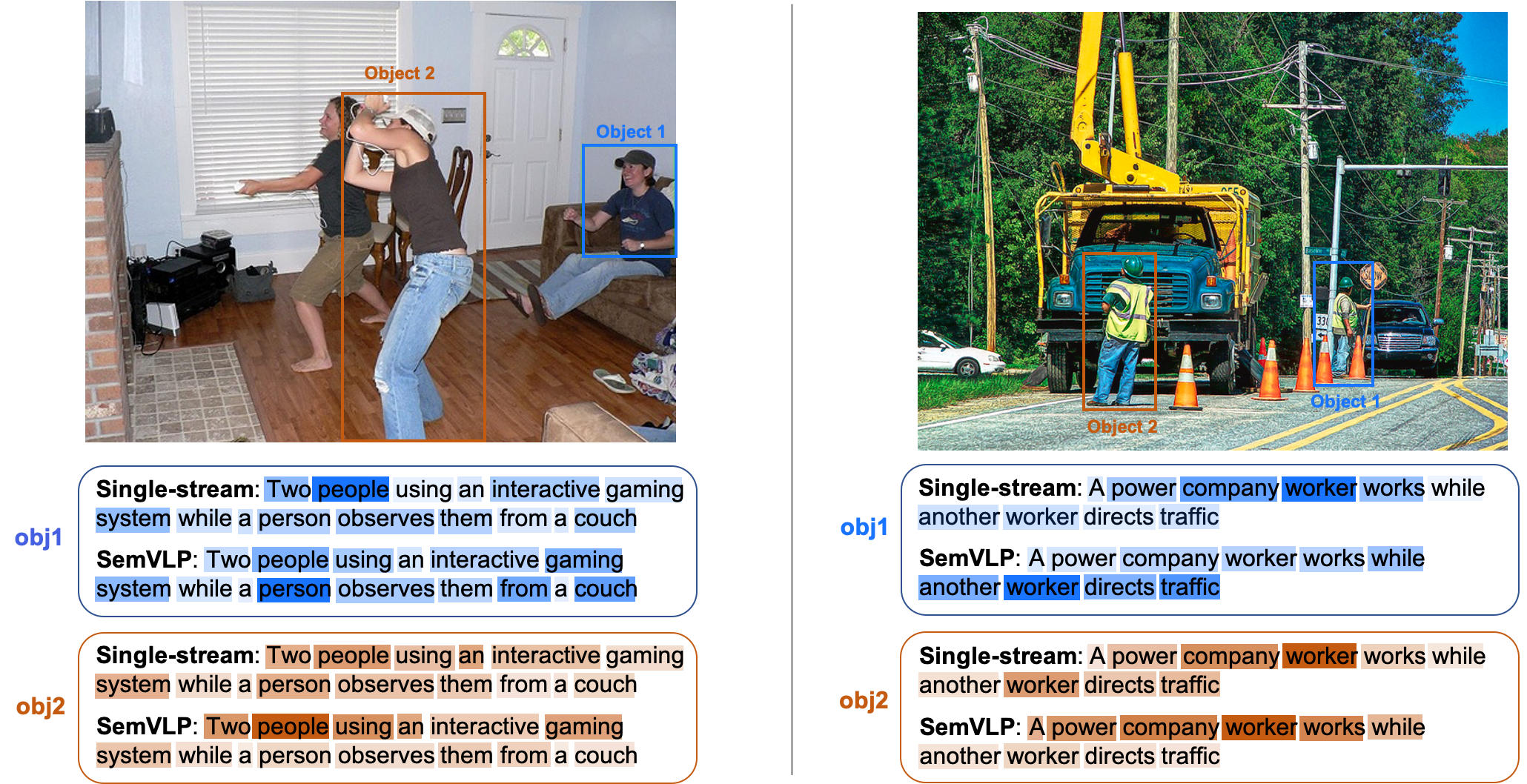}
\caption{Visualization of the Image-to-text attention example on single-stream model and SemVLP\protect~\footnotemark[3]. Object 1 and Object 2 are extracted by a pre-trained object detector Faster R-CNN.} 
\label{fig:visualization}
\end{figure*}
\footnotetext[3]{\small{We only compare the single-stream and SemVLP because the cross-attention module of two-stream model will interfere with the visualization.}}

\textbf{Analysis on Different High-level Semantic Alignments} \quad There are many different ways for high-level semantic alignment, now we further analyze the advantage of our two-stream architecture and the specific ``point'' to conduct modality fusion with the cross-modal attention module. Therefore, we pre-train SemVLP with only high-level semantic alignment and examine in which layer to introduce the cross-modal attention module by setting different $L_s$. The pre-training details are kept the same as in the original SemVLP pre-training. We test the performance on VQA and NLVR2 tasks, and the results are shown in Figure~\ref{fig:analysis2}. We can see that by introducing the cross-modal attention module properly, the two-stream mode of SemVLP method obtains significantly better performance than the previous two-stream model LXMERT.  The best performance is obtained when $L_s=6$, where the separated image/text encoding and cross-modal attention is equally emphasized. It again demonstrates the importance of aligning cross-modal representations at a proper semantic level. 

\subsection{Comparison of Fine-tuning modes}

After SemVLP is fully pre-trained at multiple semantic levels, we further examine what fine-tuning architecture/mode is more appropriate for the downstream tasks. We keep the fine-tuning setting as the same with the original setting in the supplementary material, and examine the performance using different architectures. 
Table~\ref{table:finetune} shows that on all the examined tasks, the single-stream fine-tuning architecture gives better performance than the two-stream one does. This is due to the fact that:
%The result is shown in Table~\ref{table:finetune}. From the result, it is surprising to see that among all the examined tasks, the single-stream fine-tuning architecture obtains better performances compared with the two-stream one. It may be due to that
(1) by learning cross-modal fusion from a fine-grained feature-level, the single-stream mode can capture full association across modalities from more basic semantic granularity with powerful self-attention mechanism, which originates from the success of BERT. (2) Our design of SemVLP shows more favor of the single Transformer encoder, which is well regularized when pre-training with both training modes via parameter sharing. The pre-training strategy tends to enhance the high-level semantic alignment into single-stream training process.

\subsection{Visualization}

The motivation behind SemVLP is to align cross-modal semantics at multiple levels by taking advantages of both single-stream and two-stream architectures. As stated above, single-stream and two-stream models are good at feature-level and high-level semantic alignments, respectively. To verify this, we visualize the attention map on the same layer and head of both the single-stream model and SemVLP for image objects and its associated description, as shown in Figure \ref{fig:visualization}. A darker color indicates a higher attention weight. Take the image on the left of Figure \ref{fig:visualization} as an example. In order to align the objects and the description, a model needs to capture the high-level text semantics: ``Two people'' are standing and playing games, and ``a person'' is sitting on the couch.
% In the picture on the left of Figure \ref{fig:visualization}, there are three people, and the corresponding description is “two people” and “a person”. In order to align the objects and the description, the model needs to understand the high-level text semantics. “Two people” are standing and playing games, and “a person” is sitting on the couch.
For Object 1 sitting alone, the single-stream model mis-attends to ``people'' in the description. SemVLP, on the other hand, properly pays high attention to ``person''. For Object 2 standing alongside another, our SemVLP attends to ``people'' correctly, while the single stream model fails to do so.
% it should align to the “person” in the description. We notice that single-stream model attends to “people” with higher weights, while SemVLP attends to “person” with higher weights.
%For object 2 of the example, it should align to the “people” in the description. Compared with sing-stream model, SemVLP attends to “people” with higher weights, which accurately aligns object 2 and “people”.
This example shows that the single-stream cannot differentiate between the semantics of the “person” and “people” at the high level, which leads to the false alignments. The same pattern can be observed from the image on the right of Figure \ref{fig:visualization}.

\section{Conclusion}
In this paper, we propose a new pre-training method SemVLP to learn the joint representation of vision and language. Different from existing VLP methods relying on a fixed-level semantic alignment, we introduce to align cross-modal semantics at multiple levels, by assembling a shared Transformer encoder and a pluggable cross-modal attention module in different ways. Experiment results on various downstream V+L tasks demonstrate the effectiveness of our method for understanding the diverse semantics behind the real-world image-text data.

\bibliography{naacl}

\begin{thebibliography}{28}
\expandafter\ifx\csname natexlab\endcsname\relax\def\natexlab#1{#1}\fi

\bibitem[{Antol et~al.(2015)Antol, Agrawal, Lu, Mitchell, Batra,
  Lawrence~Zitnick, and Parikh}]{antol2015vqa}
Stanislaw Antol, Aishwarya Agrawal, Jiasen Lu, Margaret Mitchell, Dhruv Batra,
  C~Lawrence~Zitnick, and Devi Parikh. 2015.
\newblock Vqa: Visual question answering.
\newblock In \emph{Proceedings of the IEEE international conference on computer
  vision}, pages 2425--2433.

\bibitem[{Chen et~al.(2019{\natexlab{a}})Chen, Gan, Li, Cheng, Wang, and
  Liu}]{chen2019meta}
Wenhu Chen, Zhe Gan, Linjie Li, Yu~Cheng, William Wang, and Jingjing Liu.
  2019{\natexlab{a}}.
\newblock Meta module network for compositional visual reasoning.
\newblock \emph{arXiv preprint arXiv:1910.03230}.

\bibitem[{Chen et~al.(2019{\natexlab{b}})Chen, Li, Yu, El~Kholy, Ahmed, Gan,
  Cheng, and Liu}]{chenuniter}
Yen-Chun Chen, Linjie Li, Licheng Yu, Ahmed El~Kholy, Faisal Ahmed, Zhe Gan,
  Yu~Cheng, and Jingjing Liu. 2019{\natexlab{b}}.
\newblock Uniter: Universal image-text representation learning.

\bibitem[{Devlin et~al.(2018)Devlin, Chang, Lee, and
  Toutanova}]{devlin2018bert}
Jacob Devlin, Ming-Wei Chang, Kenton Lee, and Kristina Toutanova. 2018.
\newblock Bert: Pre-training of deep bidirectional transformers for language
  understanding.
\newblock \emph{arXiv preprint arXiv:1810.04805}.

\bibitem[{He et~al.(2016)He, Zhang, Ren, and Sun}]{he2016deep}
Kaiming He, Xiangyu Zhang, Shaoqing Ren, and Jian Sun. 2016.
\newblock Deep residual learning for image recognition.
\newblock In \emph{Proceedings of the IEEE conference on computer vision and
  pattern recognition}, pages 770--778.

\bibitem[{Huang et~al.(2020)Huang, Zeng, Liu, Fu, and Fu}]{huang2020pixel}
Zhicheng Huang, Zhaoyang Zeng, Bei Liu, Dongmei Fu, and Jianlong Fu. 2020.
\newblock Pixel-bert: Aligning image pixels with text by deep multi-modal
  transformers.
\newblock \emph{arXiv preprint arXiv:2004.00849}.

\bibitem[{Hudson and Manning(2019{\natexlab{a}})}]{hudson2019learning}
Drew Hudson and Christopher~D Manning. 2019{\natexlab{a}}.
\newblock Learning by abstraction: The neural state machine.
\newblock In \emph{Advances in Neural Information Processing Systems}, pages
  5903--5916.

\bibitem[{Hudson and Manning(2019{\natexlab{b}})}]{hudson2019gqa}
Drew~A Hudson and Christopher~D Manning. 2019{\natexlab{b}}.
\newblock Gqa: A new dataset for real-world visual reasoning and compositional
  question answering.
\newblock In \emph{Proceedings of the IEEE Conference on Computer Vision and
  Pattern Recognition}, pages 6700--6709.

\bibitem[{Krishna et~al.(2017)Krishna, Zhu, Groth, Johnson, Hata, Kravitz,
  Chen, Kalantidis, Li, Shamma et~al.}]{krishna2017visual}
Ranjay Krishna, Yuke Zhu, Oliver Groth, Justin Johnson, Kenji Hata, Joshua
  Kravitz, Stephanie Chen, Yannis Kalantidis, Li-Jia Li, David~A Shamma, et~al.
  2017.
\newblock Visual genome: Connecting language and vision using crowdsourced
  dense image annotations.
\newblock \emph{International journal of computer vision}, 123(1):32--73.

\bibitem[{Lan et~al.(2019)Lan, Chen, Goodman, Gimpel, Sharma, and
  Soricut}]{lan2019albert}
Zhenzhong Lan, Mingda Chen, Sebastian Goodman, Kevin Gimpel, Piyush Sharma, and
  Radu Soricut. 2019.
\newblock Albert: A lite bert for self-supervised learning of language
  representations.
\newblock In \emph{International Conference on Learning Representations}.

\bibitem[{Li et~al.(2020)Li, Yin, Li, Hu, Zhang, Zhang, Wang, Hu, Dong, Wei
  et~al.}]{li2020oscar}
Xiujun Li, Xi~Yin, Chunyuan Li, Xiaowei Hu, Pengchuan Zhang, Lei Zhang, Lijuan
  Wang, Houdong Hu, Li~Dong, Furu Wei, et~al. 2020.
\newblock Oscar: Object-semantics aligned pre-training for vision-language
  tasks.
\newblock \emph{arXiv preprint arXiv:2004.06165}.

\bibitem[{Lin et~al.(2014)Lin, Maire, Belongie, Hays, Perona, Ramanan,
  Doll{\'a}r, and Zitnick}]{lin2014microsoft}
Tsung-Yi Lin, Michael Maire, Serge Belongie, James Hays, Pietro Perona, Deva
  Ramanan, Piotr Doll{\'a}r, and C~Lawrence Zitnick. 2014.
\newblock Microsoft coco: Common objects in context.
\newblock In \emph{European conference on computer vision}, pages 740--755.
  Springer.

\bibitem[{Lu et~al.(2019)Lu, Batra, Parikh, and Lee}]{lu2019vilbert}
Jiasen Lu, Dhruv Batra, Devi Parikh, and Stefan Lee. 2019.
\newblock Vilbert: Pretraining task-agnostic visiolinguistic representations
  for vision-and-language tasks.
\newblock In \emph{Advances in Neural Information Processing Systems}, pages
  13--23.

\bibitem[{Lu et~al.(2020)Lu, Goswami, Rohrbach, Parikh, and Lee}]{lu202012}
Jiasen Lu, Vedanuj Goswami, Marcus Rohrbach, Devi Parikh, and Stefan Lee. 2020.
\newblock 12-in-1: Multi-task vision and language representation learning.
\newblock In \emph{Proceedings of the IEEE/CVF Conference on Computer Vision
  and Pattern Recognition}, pages 10437--10446.

\bibitem[{Ordonez et~al.(2011)Ordonez, Kulkarni, and Berg}]{ordonez2011im2text}
Vicente Ordonez, Girish Kulkarni, and Tamara~L Berg. 2011.
\newblock Im2text: Describing images using 1 million captioned photographs.
\newblock In \emph{Advances in neural information processing systems}, pages
  1143--1151.

\bibitem[{Radford et~al.(2018)Radford, Narasimhan, Salimans, and
  Sutskever}]{radford2018improving}
Alec Radford, Karthik Narasimhan, Tim Salimans, and Ilya Sutskever. 2018.
\newblock Improving language understanding by generative pre-training.
\newblock \emph{URL https://s3-us-west-2. amazonaws.
  com/openai-assets/researchcovers/languageunsupervised/language understanding
  paper. pdf}.

\bibitem[{Raffel et~al.(2019)Raffel, Shazeer, Roberts, Lee, Narang, Matena,
  Zhou, Li, and Liu}]{raffel2019exploring}
Colin Raffel, Noam Shazeer, Adam Roberts, Katherine Lee, Sharan Narang, Michael
  Matena, Yanqi Zhou, Wei Li, and Peter~J Liu. 2019.
\newblock Exploring the limits of transfer learning with a unified text-to-text
  transformer.
\newblock \emph{arXiv preprint arXiv:1910.10683}.

\bibitem[{Ren et~al.(2015)Ren, He, Girshick, and Sun}]{ren2015faster}
Shaoqing Ren, Kaiming He, Ross Girshick, and Jian Sun. 2015.
\newblock Faster r-cnn: Towards real-time object detection with region proposal
  networks.
\newblock In \emph{Advances in neural information processing systems}, pages
  91--99.

\bibitem[{Sharma et~al.(2018)Sharma, Ding, Goodman, and
  Soricut}]{sharma2018conceptual}
Piyush Sharma, Nan Ding, Sebastian Goodman, and Radu Soricut. 2018.
\newblock Conceptual captions: A cleaned, hypernymed, image alt-text dataset
  for automatic image captioning.
\newblock In \emph{Proceedings of the 56th Annual Meeting of the Association
  for Computational Linguistics (Volume 1: Long Papers)}, pages 2556--2565.

\bibitem[{Su et~al.(2019)Su, Zhu, Cao, Li, Lu, Wei, and Dai}]{su2019vl}
Weijie Su, Xizhou Zhu, Yue Cao, Bin Li, Lewei Lu, Furu Wei, and Jifeng Dai.
  2019.
\newblock Vl-bert: Pre-training of generic visual-linguistic representations.
\newblock \emph{arXiv preprint arXiv:1908.08530}.

\bibitem[{Suhr et~al.(2018)Suhr, Zhou, Zhang, Zhang, Bai, and
  Artzi}]{suhr2018corpus}
Alane Suhr, Stephanie Zhou, Ally Zhang, Iris Zhang, Huajun Bai, and Yoav Artzi.
  2018.
\newblock A corpus for reasoning about natural language grounded in
  photographs.
\newblock \emph{arXiv preprint arXiv:1811.00491}.

\bibitem[{Tan and Bansal(2019)}]{tan2019lxmert}
Hao Tan and Mohit Bansal. 2019.
\newblock Lxmert: Learning cross-modality encoder representations from
  transformers.
\newblock \emph{arXiv preprint arXiv:1908.07490}.

\bibitem[{Vaswani et~al.(2017)Vaswani, Shazeer, Parmar, Uszkoreit, Jones,
  Gomez, Kaiser, and Polosukhin}]{vaswani2017attention}
Ashish Vaswani, Noam Shazeer, Niki Parmar, Jakob Uszkoreit, Llion Jones,
  Aidan~N Gomez, {\L}ukasz Kaiser, and Illia Polosukhin. 2017.
\newblock Attention is all you need.
\newblock In \emph{Advances in neural information processing systems}, pages
  5998--6008.

\bibitem[{Wang et~al.(2019)Wang, Bi, Yan, Wu, Bao, Xia, Peng, and
  Si}]{wang2019structbert}
Wei Wang, Bin Bi, Ming Yan, Chen Wu, Zuyi Bao, Jiangnan Xia, Liwei Peng, and
  Luo Si. 2019.
\newblock Structbert: Incorporating language structures into pre-training for
  deep language understanding.
\newblock \emph{arXiv preprint arXiv:1908.04577}.

\bibitem[{Xia et~al.(2019)Xia, He, Tan, Tian, He, and Qin}]{xia2019tied}
Yingce Xia, Tianyu He, Xu~Tan, Fei Tian, Di~He, and Tao Qin. 2019.
\newblock Tied transformers: Neural machine translation with shared encoder and
  decoder.
\newblock In \emph{Proceedings of the AAAI Conference on Artificial
  Intelligence}, volume~33, pages 5466--5473.

\bibitem[{Young et~al.(2014)Young, Lai, Hodosh, and
  Hockenmaier}]{young2014image}
Peter Young, Alice Lai, Micah Hodosh, and Julia Hockenmaier. 2014.
\newblock From image descriptions to visual denotations: New similarity metrics
  for semantic inference over event descriptions.
\newblock \emph{Transactions of the Association for Computational Linguistics},
  2:67--78.

\bibitem[{Yu et~al.(2020)Yu, Tang, Yin, Sun, Tian, Wu, and Wang}]{yu2020ernie}
Fei Yu, Jiji Tang, Weichong Yin, Yu~Sun, Hao Tian, Hua Wu, and Haifeng Wang.
  2020.
\newblock Ernie-vil: Knowledge enhanced vision-language representations through
  scene graph.
\newblock \emph{arXiv preprint arXiv:2006.16934}.

\bibitem[{Zhu et~al.(2016)Zhu, Groth, Bernstein, and Fei-Fei}]{zhu2016visual7w}
Yuke Zhu, Oliver Groth, Michael Bernstein, and Li~Fei-Fei. 2016.
\newblock Visual7w: Grounded question answering in images.
\newblock In \emph{Proceedings of the IEEE conference on computer vision and
  pattern recognition}, pages 4995--5004.

\end{thebibliography}
\bibliographystyle{acl_natbib}

\appendix

%\section{Example Appendix}
%\label{sec:appendix}
%This is an appendix.

\end{document}